\documentclass{article}

\usepackage{graphicx}
\usepackage{amssymb, amsfonts, amsmath}
\usepackage{xfrac}
\usepackage{subfigure}
\usepackage{url}
\usepackage{ulem}
\usepackage{chngcntr}
\usepackage{float}
\usepackage{multirow}
\restylefloat{table}
\usepackage{color}\definecolor{darkblue}{rgb}{0,0,0.75}\definecolor{darkred}{rgb}{0.51,0.02,0}
\usepackage[pagebackref, breaklinks,colorlinks=true, citecolor=darkblue, urlcolor=darkblue,linkcolor=darkblue]{hyperref}

\usepackage{algorithm} 
\usepackage{algorithmic}  
\usepackage[algo2e]{algorithm2e} 

\usepackage{wrapfig}
\usepackage{footnote}
\usepackage{microtype}
\usepackage{arydshln}

\makesavenoteenv{tabular}
\makesavenoteenv{table}

\newcommand{\N}{\mathbb{N}}
\newcommand{\D}{\mathcal{D}}

\newcommand{\ourmaintitle}{Informed Novelty Detection in Sequential Data by Per-Cluster Modeling}
\newcommand{\ourtitle}{\ourmaintitle}

\usepackage[accepted]{icml2023}

\icmltitlerunning{\ourtitle}

\begin{document}

    \twocolumn[
        \icmltitle{\ourtitle}

        \begin{icmlauthorlist}
        \icmlauthor{Linara Adilova}{rub}
        \icmlauthor{Siming Chen}{fudan}
        \icmlauthor{Michael Kamp}{ikim,rub,monash}
        
        \end{icmlauthorlist}
        
        \icmlaffiliation{ikim}{Institute for AI in medicine (IKIM) at University Hospital Essen, Essen, Germany}
        \icmlaffiliation{rub}{Ruhr-University Bochum, Bochum, Germany}
        \icmlaffiliation{fudan}{School of Data Science, Fudan University, Shanghai, China}
        \icmlaffiliation{monash}{Monash University, Melbourne Australia}
        
        \icmlcorrespondingauthor{Linara Adilova}{linara.adilova@rub.de}

        \icmlkeywords{novelty detection, visual analytics, human-in-the-loop, decomposition, clustering, sequence modeling}
        
        \vskip 0.3in
    ]

\printAffiliationsAndNotice{}

\begin{abstract}
Novelty detection in discrete sequences is a challenging task, since deviations from the process generating the normal data are often small or intentionally hidden. 
%Novelties can be detected by modeling normal sequences and measuring the deviations of a new sequence from the model predictions. 
%Such models can be trained on a set of normal sequences in an unsupervised manner, i.e., without a novelty label. 
%However, 
In many applications data is generated by several distinct processes
%(e.g., different behaviors in interactions with a web service) 
so that models trained on all the data tend to over-generalize and novelties remain undetected.
We propose to approach this challenge through decomposition: by clustering the data we break down the problem, obtaining simpler modeling tasks in each cluster which can be modeled more accurately. However, this comes at a cost, since the amount of training data per cluster is reduced. This is a particular problem for discrete sequences where state-of-the-art models are data-hungry.
The success of this approach thus depends on the quality of the clustering, i.e., whether the individual learning problems are sufficiently simpler than the joint problem. 
%While clustering discrete sequences automatically is a challenging and domain-specific task, it is often easy for human domain experts, given the right tools. 
%In this paper, we adapt a state-of-the-art visual analytics tool for discrete sequence clustering to obtain informed clusters from domain experts and use LSTMs to model each cluster individually. 
In this paper we adapt a state-of-the-art visual analytics tool for discrete sequence clustering to obtain informed clusters from domain experts, since clustering discrete sequences automatically is a challenging and domain-specific task.
We use LSTMs to further model each of the clusters.
Our empirical evaluation indicates that this informed clustering outperforms automatic ones and that our approach outperforms standard novelty detection methods for discrete sequences in three real-world application scenarios.
%In particular, decomposition outperforms a global model despite less training data on each individual cluster.
\end{abstract}

\section{Introduction}
\label{sec:introduction}

The task of identifying point anomalies, as classified by \cite{chandola2009anomaly}, with respect to previously observed data is at the core of many applications: in cybersecurity, a novel kind of interaction with a web service can indicate an attack, and in news verification, detecting a deviation in the writing style can hint at an article being fake. If a dataset of previously observed instances without anomalies is available, this form of anomaly detection is called novelty detection. This task is particularly challenging for discrete sequential data, since deviations may occur only in the order of elements or in the frequency of patterns within the sequence. Usually anomaly and novelty detection methods are based on some form of similarity measure between instances, but approaches that only compare sequences by the elements that they contain will fail in such cases. Instead, the state-of-the-art approach is to use metrics designed for sequences, such as the longest common sub-sequence distance~\citep{chandola2008comparative, budalakoti2006anomaly}, or to use sequential pattern mining to extract features to be used with Euclidean distances~\citep{feremans2019pattern}. Nevertheless, it is always hard to choose the features that will suit the task at hand or the distance measure that captures the novel behavior best.
A more general approach is to model the process that generates the sequences~\citep{marchi2015novel, warrender1999detecting, florez2005efficient}
and check whether an observed sequence is likely under that model. For this predictive modeling to work, it is paramount that the process is imitated accurately, since novelties often deviate only slightly from previously observed sequences. Recurrent neural networks, such as long short-term memory networks (LSTMs) or gated recurrent unit networks (GRUs), achieve high accuracy, outperforming both sequential metrics and pattern mining approaches~\citep{chalapathy2019deep}. However, often sequences are generated not by a single process, but by several distinct processes, e.g., different behavioral patterns of users of a system or separate topics of news that will vary in the domain-specific language used.
Modeling all of them jointly can lead to over-generalization which results in a lower sensitivity to small deviations. 
In predictive modeling, an approach to improve the modeling accuracy is decomposition: the data is decomposed into parts that are supposed to constitute easier sub-problems. The idea is that models trained on each part---sometimes referred to as local experts~\citep{nowlan1991evaluation}---outperform the global model~\citep{sharkey1999linear}. Even though this decomposition reduces the available training data per model, it has been observed that given a good decomposition, this approach is beneficial. For example, in natural language processing, domain-specific modeling outperforms global models, even for high-capacity model classes such as recurrent neural networks~\citep{joshi2012multi}. 
%
% It is striking of course, that reducing amount of training data might lead to accuracy improvement, but in the case of truly simplified learning task, i.e., clusters that describe a simpler task, the accuracy can be significantly improved even with less data. Since the adoption of LSTMs for sequence learning is equvivalent by learning task to language modeling in natural language processing area, we can conclude that domain features knowledge and data separation is very beneficial~\citep{joshi2012multi}. 
%

This paper proposes to use decomposition in novelty detection in discrete sequences. 
%based on human knowledge to obtain more accurate models for novelty detection in discrete sequences. 
%
\begin{figure}[ht]
\centering
  \includegraphics[width=0.7\columnwidth]{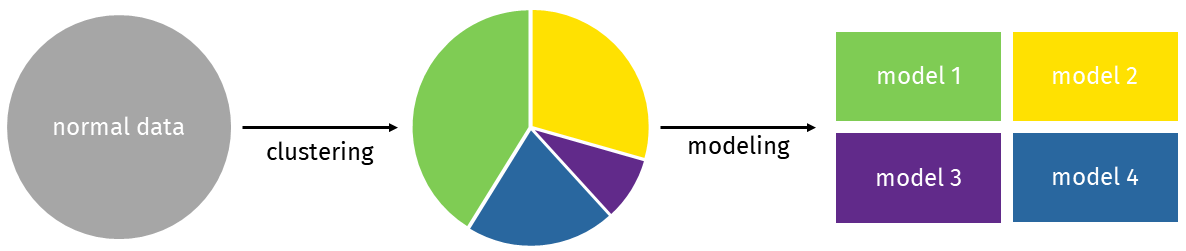}
  \caption{Schematic illustration of the training phase.}
  \label{fig:schema_training}
\end{figure}
\begin{figure}[ht]
\centering
  \includegraphics[width=0.7\columnwidth]{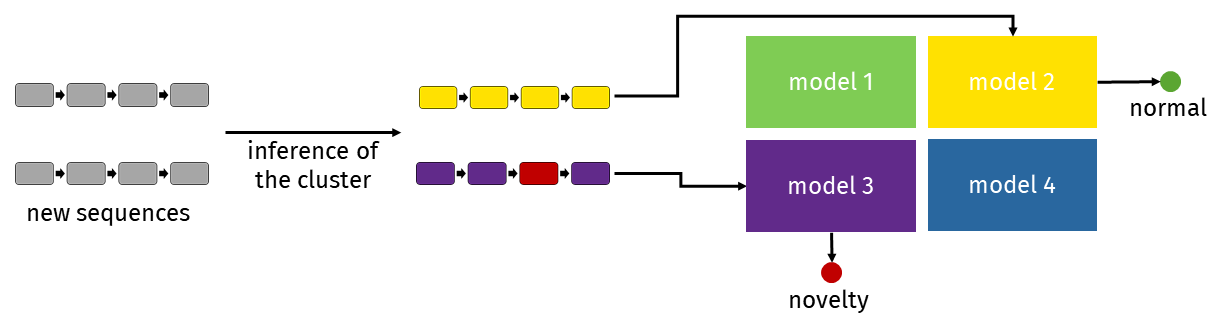}
  \caption{Schematic illustration of the inference phase.}
  \label{fig:schema_inference}
\end{figure}
That is, the training data is decomposed into clusters, on each of which a model is trained (Fig.~\ref{fig:schema_training}). During inference, a new data point is assigned to a cluster and the likelihood under the respective model is used to determine whether it is a novelty (Fig.~\ref{fig:schema_inference}). This general framework allows for many choices of decomposition and modeling techniques. 
A major challenge is identifying the right decomposition automatically. This is hard for sequential data, in particular, due to its high dimensionality and structure. However, for a human it is often quite simple to find meaningful groups in data. Visual analytics tools for including a human in the loop are widely employed directly for anomaly detection~\citep{janetzko2014anomaly, leite2018visual} and also for analysis of behavior as sequences of actions in various applications~\citep{shi2020visual}.
%For example in cybersecurity, security operators track suspicious activities of users in a system and intuitively know the typical behavior patterns of certain groups of users. Similarly, in fake review detection, a human expert can group texts based on application-dependent topics.  
%
This paper uses visual analytics to obtain a human knowledge based, semantically informed clustering. For that, we extend a visual analytics tool developed for finding user behavior clusters~\citep{chen2019lda}. This interface was evaluated by domain experts and got very positive feedback as a clustering technique allowing for deep insights into the sequential data and flexible decomposition of sequences into semantically sound clusters\footnote{For a detailed demonstration see \url{http://simingchen.me/docs/tvcg19_lda.mp4}}. We derive a method for assigning new incoming sequences to clusters obtained by this tool in order to use it in our framework. 
We then train LSTMs~\citep{lstm1997hochreiter} per cluster and use them to determine whether a new sequence is a novelty based on the perplexity score of their predictions---a widely applied technique for evaluating the quality of language models in natural language processing application~\citep{jurafsky2000speech}.
%The sequences are modeled then with LSTMs~\citep{lstm1997hochreiter}.  
% That is, sequences are modeled by LSTMs~\citep{lstm1997hochreiter} trained on partitions of the data obtained from a human expert informed clustering.
%In order to decide whether a sequence is a novelty, we obtain the perplexity score from the LSTM corresponding to its assigned cluster. If perplexity---a widely applied technique for evaluating the quality of language models~\citep{jurafsky2000speech}---surpasses a given threshold, the sequence is identified as a novelty. 
%

This decomposition framework for novelty detection in discrete sequences differs fundamentally from the classical use of clustering to identify anomalies directly~\citep{pavlov2003sequence, chandola2008comparative, cadez2000visualization}. It also differs from approaches that use modeling only to obtain a representation of the data, e.g., the use of the transformation by an autoencoder neural network as feature vector for classical anomaly detection methods~\citep{yuan2017deep,corizzo2020scalable}.  % example, the internal representation of the data by a neural network, like an autoencoder, can be used as feature vector for a classical anomaly detection method~\citep{yuan2017deep,corizzo2020scalable}. 
We show that this approach outperforms a global model trained on all data in three real-world applications, despite the fact that the amount of training data for the global model is substantially larger than for the cluster models. We also show that the human-knowledge based informed clustering can outperform automatic clustering techniques, e.g., k-nearest neighbor (kNN) or Latent Dirichlet Allocation (LDA). Moreover, this approach substantially outperforms standard approaches for novelty detection in discrete sequential data based on sequence metrics and sequential pattern mining. Thus, this technique can improve the detection of fake news or attacks in cybersecurity.

In summary, the contributions are: (i) a novel framework that combines informed decomposition and modeling for novelty detection in discrete sequential data, (ii) an extension of a visual analytics tool that allows human experts to easily identify meaningful clusters of discrete sequences, and (iii) an evaluation of the framework for novelty detection in sequential data in three real-world scenarios. 
\section{Related Work}
\label{sec:related_work}

\subsection{Novelty Detection in Discrete Sequences}
\label{sec:novelty_discrete_sequences}

Novelty and anomaly detection are particularly challenging for discrete sequences~\citep{domingues2019comparative, chandola2010anomaly}, since anomalies often only deviate in the order of elements or the frequency of patterns. State-of-the-art approaches use common outliers detection algorithms in combination with sequence specific metrics, or sequence specific features (e.g., sequential patterns). However, these approaches are computationally expensive and become infeasible for large amounts of data and long sequences: computing the longest common sub-sequence of two sequences has a runtime linear in the product of their lengths, and sequential pattern mining is in \#P~\citep{dong2007sequence}.
Note that detecting novel discrete sequences is fundamentally different from outlier detection in time-series, where a single time point constitutes an outlier. Therefore, approaches for outlier detection in time series~\citep[][cf.]{ren2019time, zhang2019deep} cannot be straight-forwardly applied.

% It should be noted here, that the task of finding sequences outliers is significantly different from the task of time-series outlier detection, where a lot of methods and approaches are proposed (e.g., \cite{ren2019time, zhang2019deep}). Time-series anomaly detection requires to identify an outlying timestep in the ongoing series of steps, while we consider a set of finite sequences and aim to identify if such sequence is an outlier compared to others. Moreover, we put attention at discrete sequences (though in the Sec.~\ref{sec:empirical_evaluation} we demonstrate that it is possible to represent a numeric sequence as discrete as well), which adds one more complexity level for modeling compared to more wide-spread numerical sequences, which allow for more feature extraction approaches.

A different way is to model the normal sequences and use the models for novelty detection. A successful approach to discrete sequence modeling is adapting tools from natural language processing, particularly LSTM-based neural language models~\citep{chalapathy2019deep}. The task of a language model is to identify the likelihood of a sequence of words and it is usually trained in an unsupervised way on a vast amount of unlabeled data, therefore it allows to understand which sequences do not belong to the learned language distribution. In the general case of discrete sequences, objects organized in sequences are treated as words, and the model learns the probability distribution over the space of object sequences. Language modeling was already employed in anomaly detection for network security. For example,~\cite{tuor2018recurrent} applied character level language modeling for separate lines of logging files in order to identify fraudulent actions. This approach was shown to perform well on one of the publicly available datasets, but is rather sensitive to the format of logging information, e.g., having information about the IP address of the request and success of the performed action. It also does not employ the information of the sequences of actions performed in each interaction. \cite{kim2016lstm} instead use an ensemble of language models that are learning normal user behavior from sequences of actions. Both \cite{tuor2018recurrent} and~\cite{kim2016lstm} mention the need of separate modeling for particular groups of behaviors. However,~\cite{tuor2018recurrent} only partitions data based on the timespan of sessions and~\cite{kim2016lstm} does not partition data at all, but instead uses an ensemble of different models. Our goal is to demonstrate that modeling applied after clustering is more beneficial than straightforward modeling or other non-modeling based approaches. The model (here LSTMs) can be replaced with specialized approaches, based e.g., on GRUs or VAEs~ \citep{su2019robust}. 

\subsection{Decomposition for Modeling}
\label{sec:decomposition}
This paper proposes to decompose data by informed clustering and model each cluster separately. 
Clustering has previously been used directly for outlier detection, either performed in the space of items themselves~\citep{campello2013density}, in the space of models~\citep{cadez2000visualization}, or in the space of features~\citep{liu2008isolation}. It groups data points together based on a notion of density, or connectivity. Outliers are data points that cannot be assigned to any cluster. However, these methods are oblivious to the modeling task, the goal is only to get clusters directly pointing to anomalies.
At the same time in data mining and knowledge extraction from data, decomposition has been empirically shown to be beneficial for modeling unlabeled interesting features~\citep{baxt1990use, buntine1996graphical}. Indeed, \cite{sharkey1999multi} argues that the decomposition is mainly motivated by the performance improvement gained through a better bias-variance trade-off for the models. 

Our paper considers decomposition in the space of items (i.e., sequences). In the literature~\citep[cf.][]{maimon2005decomposition} this is termed space decomposition or horizontal decomposition. Examples are mixture of experts~\citep{nowlan1991evaluation}, local linear regression~\citep{draper1981applied}, or adaptive subspace models~\citep{ramamurti1999structurally}. Such decomposition is performed during training, while our approach is to cluster beforehand. 
Up to our knowledge, decomposition has not yet been used for novelty detection by modeling the data generating process. This proposed approach differs from previous methods, such as local modeling and ensembles, in that it decouples decomposition from modeling and thus is more likely to avoid overfitting. The approach is similar in spirit to the work of~\cite{bergman2019classification}, where multiple transformations of a dataset are created and an outlier detection model is learned for each transformation, but aims at finding such subspaces within the initial space itself.

Note that the additional runtime due to decomposition is often negligible, since it is performed only once, while individual models in this case process less training data which can actually improve the runtime. Moreover, in some applications, lower-capacity models can be chosen as local models to further improve the runtime. 

In this work, we choose high-capacity models for each cluster for which it might be an issue that decomposition leads to parts of the data having so few training examples that no meaningful model will be learned. 
Surprisingly, our empirical evaluation shows that for the task of novelty detection well-performing models can be trained even on very limited amounts of data, depending on the clustering method. We conjecture that in this case the decomposition through clustering indeed reduces the complexity of the learning task, and that for novelty detection slight overfitting on a local dataset is not as detrimental as in predictive modeling. 
Moreover, in many applications data is abundant so that local cluster size will not be an issue. For example, in cybersecurity \cite{faraoun2006neural} note that the amount of data is constantly growing, e.g., log files of network monitoring systems are constantly updated. 

%In this case, partitioning the data is even beneficial because training a model on all data becomes infeasible.
%For example, \cite{chandrasekhar2014confederation} employed fuzzy clustering to separate data into homogeneous subsets and learn separate artificial neural networks for each of the subsets for further predictions. But these approaches do not allow to employ sequential information contained in the sessions. Also these works are considering intrusion detection as a supervised task with possibly inbalanced training set.

It should be noted that for sequence modeling, where sequences are generated from different processes, a global model trained on the entire data may perform poorly, since it is difficult to identify these processes automatically (unless additional information, such as labels, is given)~\citep{joshi2012multi}.
Similarly, automatic decomposition would be a challenging task.
%At the same time, decomposing the data with respect to those processes with automatic methods is a challenging task.
However, for human experts it is often easy to cluster data. Visual analytics provides a natural interface that allows to extract this implicit and intuitive knowledge of human experts~\citep{liu2017towards}.
\section{Combining Clustering and Novelty Detection}
\label{sec:formal_description}
We assume a dataset $X$ of $n\in\N$ normal sequences $s\in \mathcal{V}^{*}$ over a finite vocabulary $\mathcal{V}$ of words, i.e., $s=(v_1,\dots,v_{\mathbf{len}(s)})$ with $v_i\in\mathcal{V}$. We assume the sequences $s$ are drawn iid. according to a distribution $\D$ over $\mathcal{V}^{*}$. For this distribution, we assume that it is a mixture of $m\in\N$ distributions $\D_1,\dots,\D_m$, where each $\D_i$ corresponds to a different process generating the sequences. For example, each $\D_i$ can represent different types of user behavior in the interaction with a web service, or different news topics. The task is to decide for a new sequence $s^\prime\in\mathcal{V}^{*}$, $s^\prime \not\in X$ if it is an anomaly with respect to $\D$, i.e., if it is a novelty.

A sequence clustering is a function $C:\mathcal{V}^{*}\rightarrow [k]$ that assigns each sequence to one of $k\in\mathbb{N}$ clusters. We assume clustering algorithms that take as input a dataset $X$ and the parameter $k$ and output a clustering $C$. We further assume that each data point has a unique cluster assigned to it---that is, the clustering is a partition of $X$ into sets $\mathcal{G}_1,\dots,\mathcal{G}_k\subseteq X$, i.e., $\bigcup_{i=1}^{k} \mathcal{G}_i = X$ and for all $i,j\in [k]$ it holds that $\mathcal{G}_i\cap \mathcal{G}_j=\emptyset$. Ideally, $k=m$ and each cluster $i$ corresponds to a distribution $\D_i$ in the mixture. 

Novelty detection on sequences is an unsupervised learning task. A powerful modelling technique for sequences are LSTM-based neural networks that model the process that generates the sequences. 
This, however, is a supervised task: given a prefix of a sequence the model $h: \mathcal{V}* \rightarrow [0,1]^{\mathbf{card}(\mathcal{V})}$ predicts the next element in a sequence by assigning each possible element from $\mathcal{V}$ a likelihood score.
That is, given the prefix $p_i=(v_1,\dots,v_{i})$ of a sequence $s=(v_1,\dots,v_{\mathbf{len}(s)})\in\mathcal{V}*$, where $i\in [\mathbf{len}(s)]$, the model outputs the likelihoods $h(p_i)\in [0,1]^{\mathbf{card}(\mathcal{V})}$.
The predicted next element is the one with the highest likelihood. 
The true label for training is a vector with a likelihood of $1$ assigned to the correct element at that place and $0$ to all other elements.
Note that an LSTM can predict further elements $j>i+1$ in the sequence, but we do not make use of this, here.  
After training on a given set of normal sequences, we can use such a process model for novelty detection: instead of using the predicted elements, we compute for each prefix $p_i$ the probability the model $h$ assigns to the actual element $v_{i+1}$ in $s$, i.e., $h(p_i)_{v_{i+1}}$. 
If one or multiple elements in a sequence are predicted to be very unlikely given their prefix, then these elements are unusual given the modeled distribution. We combine the individual probabilities into a novelty score using the perplexity score~\citep{jurafsky2000speech}
\begin{equation} \label{eq:perplexity}
PP(h,s) := \sqrt[\leftroot{-3}\uproot{3}{\mathbf{len}(s)}]{\prod_{i=1}^{\mathbf{len}(s)}\frac{1}{h(p_i)_{v_{i+1}}}}
\end{equation}
where $\mathbf{len}(s)$ is the number of words in a sequence and $h(p_i)_{v_{i+1}}$ is the predicted likelihood of $v_{i+1}$, the $(i+1)$-th element in $s$. The larger the perplexity score, the less probable it is that the sequence has originated from the same distribution that the model was trained on. Thus, a high perplexity score indicates novelty.

\begin{algorithm}
\caption{Novelty Detection via Per-Cluster Modelling}\label{algo:framework}
\hspace*{\algorithmicindent} \textbf{Input} dataset $X\subset {\mathcal{V}^{*}}$, threshold $\theta\in\mathbb{R}$, sequence $s'\in\mathcal{V}^*$\\
\hspace*{\algorithmicindent} \textbf{Output} $\{0,1\}$  ($0$ for a normal sequence, $1$ for a novelty)
\begin{algorithmic}[1]
  \STATE \textbf{Training:}
  \STATE  obtain clustering $C$ with $k$ clusters of $X$ %with parameter $k$ optimized via silhouette values 
  \FOR{$\mathcal{G}_i=\{s\in X \mid C(s)=i\}$ with $i=1,\dots,k$}
    \STATE  train process model $h_i$ on $\mathcal{G}_i$
  \ENDFOR
  \STATE  \textbf{Inference:}
  \STATE  compute cluster $C(s')$ of $s'$
  \IF{$PP(h_{C(s')},s') > \theta$}
      \RETURN  $1$
  \ELSE
      \RETURN  $0$
  \ENDIF
\end{algorithmic}
\end{algorithm}

% \begin{algorithm*}
% \caption{Novelty Detection via Per-Cluster Modelling}\label{algo:framework}
% \hspace*{\algorithmicindent} \textbf{Input} dataset $X\subset {\mathcal{V}^{*}}^n$, threshold $\theta\in\mathbb{R}$, sequence $s\in\mathcal{V}^*$\\
% \hspace*{\algorithmicindent} \textbf{Output} $\{0,1\}$
% \begin{algorithmic}%[1]
% \COMMENT{Training:}
% \STATE obtain clustering $C$ of $X$ with parameter $k$ optimized via silhouette values 
% \FOREACH{$\mathcal{G}_i\subseteq X$ with $i\in [k]$}
%     \STATE train process model $h_i$ on $\mathcal{G}_i$
% \ENDFOR
% \COMMENT{Inference:}
% \State compute cluster $C(s)$ of $s$
% \If{$PP(h_{C(s)},s) > \theta$}
%     \RETURN  output $1$, i.e., $s$ is a novelty
% \Else
%     \RETURN  output $0$, i.e., $s$ is not a novelty
% \EndIf
% \end{algorithmic}
% \end{algorithm*}
%
Modelling on a sample $X$ from $\D$ requires the model to generalize over all $m$ distributions $\D_i$ in the mixture which can result in lower sensitivity to small deviations. Ideally, we want to train an individual model for each $\D_i$. Since the actual mixture is typically unknown, we use the clustering as an approximation. The proposed approach, given in Algorithm~\ref{algo:framework}, first using a clustering algorithm on the input data $X$ to obtain a clustering $C$. The optimal number of clusters $k$ can be determined via the standard silhouette value~\citep{rousseeuw1987silhouettes}. Then, a model $h_i$ is trained for each cluster $i$. The novelty score of a sequence $s'$ is obtained by first assigning $s'$ to a cluster $C(s')$ and then computing $PP(h_{C(s')},s')$. If $PP(h_{C(s')},s')>\theta$, $s'$ is reported as novelty.

As described above, this approach follows the intuition that a model trained for each individual distribution $\D_i$ in the mixture is more precise for that distribution, such that novelties will be detected more accurately. Given a good clustering, i.e., one for which clusters correspond to distributions in the mixture, the novelty detection capabilities of the models should improve over a model trained on the entire mixture. At the same time, the clustering reduces the effective training set size for the individual models, in turn reducing their capabilities. In Section~\ref{sec:empirical_evaluation} we show empirically on a variety of different applications that the advantage of per-cluster modelling substantially outweighs the reduction in training set size. Since the success of this approach highly depends on the quality of clustering, we propose to use a visual analytics based approach for human experts to inform an LDA-based clustering in the following section.

\section{Visual Informed Clustering}
\label{sec:informed_clustering}
%The injection of knowledge is challenging in the context of cybersecurity. For example, \cite{meng2015design} involved experts knowledge in the loop for recognizing the most successful models for reducing the amount of false alarms, at the same time keeping the level of accuracy high. But it is usually hard to provide interpretable information for getting valuable input from experts \citep{sommer2010outside}, while our approach allows for intuitive visual interaction to achieve this goal.
To obtain a good decomposition of sequences, the input data has to be clustered in a meaningful way. Techniques like k-means or topic modeling with LDA are able to cluster sequences, but often fail to find application-domain specific partitions and to be meaningful for high-dimensional data. %Moreover, these methods require to set the number of expected clusters or topics a priori, which is challenging in itself~\cite{Liu2016AnOO,Mimno:2007:MHT}.
%COMMENT: In the new version, we say in our method that we "easily" find the right number of clusters via silhouette values. Thus, I would avoid pointing this out.
Experts often have domain knowledge that can improve clustering, but it is non-trivial to extract this knowledge.
We adapt a visual analytics tool that is proven to enable domain experts to cluster data in an intuitive and comfortable way~\citep{chen2019lda}. With this tool, semantically meaningful \textit{informed clusters} can be defined without setting any parameters a priori. The interactive visual analytics tool was introduced by \cite{chen2019lda} for behavior clustering. It was thoroughly evaluated in case studies with security management system and amusement park visiting behaviors. The domain expert's feedback confirmed the usefulness and efficiency of the visual approach for identifying meaningful groups of behaviors. Based on this positive evaluation, we incorporate the tool into the novelty detection framework and extend it with an inference technique for identifying the cluster of a new sequence based on the analysis performed by experts.
The tool performs multiple topic modeling runs on the normal data $X=\{s_1,\dots,s_n\}$. For that, each of the sequences is transformed into a bag of words: each sequence $s$ is represented by a vector $b\in\mathbb{N}^{\mathbf{card}(\mathcal{V})}$, where $b_i$ is the count of word $w_i\in\mathcal{V}$ in $s$. In the following we refer to any kind of discrete objects organized into sequences as words (for consistensy with NLP terminology). Then, multiple rounds of LDA~\citep{blei2003latent}, each with a different setting of the number of topics, are used to produce a set of topics $T=\{t_1,...,t_{\mathbf{card}(T)}\}$. 
%The LDA models are saved for further inference of the clusters for new coming sequences. 
From that we produce an initial visualization based on a topic-word matrix (i.e., probabilities corresponding to each of the words in $\mathcal{V}$ for each of the topics in $T$), and a sequence-topic matrix (i.e., probabilities corresponding to each of the topics in $T$ according to the corresponding LDA model for each of the sequences from $X$).% are produced as a base of the initial visualization in the interface. 
% The distance between two topics $t_{k_1}, t_{k_2} \in T$ is defined as the accumulated differences of the probabilities of actions in the topics $\sqrt (\sum_{l=1}^{d}(M^{TA}_{k_{1}l} - M^{TA}_{k_{2}l}))$. 
%

\begin{figure}[!htb] 
  \centering 
  \includegraphics[width=0.62\columnwidth]{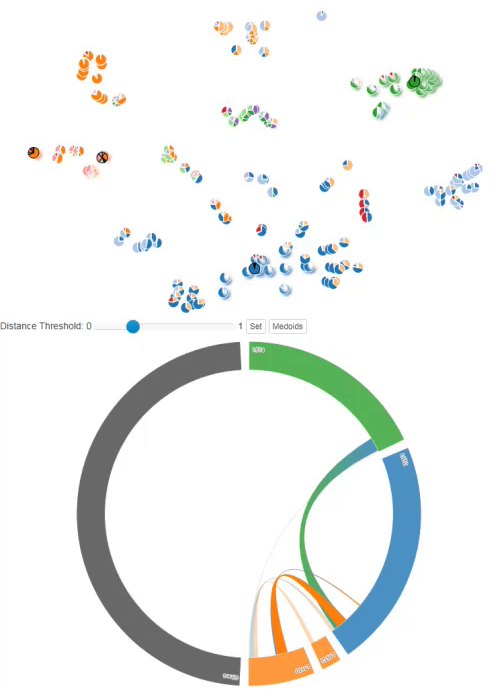} 
\caption{Partial exemplary view of the visual interface for the domain experts for understanding the distribution of data, creating and exploring properties of clusters. The topics with highlighted strokes are the medians of the informed clusters. In this case, $3$ clusters were created (leftover examples in gray glyph).
%Three clusters are already identified and the fourth one is being investigated.
%Exemplary view of the visual interface for the domain experts for understanding the distribution of data, creating and exploring clusters. The topics with highlighted strokes are the medians of the clusters created by experts. In this case, the experts created $13$ clusters.
}
  \label{fig:vis} 
\end{figure}

%The visual analytics tool's interface works as follows.
On startup, the interface displays the initial visualization based on the topic-word and sequence-topic matrix obtained from the multiple runs of LDA. %Fig.~\ref{fig:vis} shows the user interface. 
%These initial topics serve as a basis for the informed clusters.
%There are two motivations for running the multiple rounds of LDA. First, although there are non-parametric topic modeling approaches like hierarchical LDA~\cite{BleiGJT2004Hierarchical}, there is no universal approach that can determine the most suitable number of topics for all types of data~\cite{Liu2016AnOO,Mimno:2007:MHT}. Second, depending on different goals security experts can 
The tool projects the generated topics (as vectors of word probabilities) on $2D$ space using t-SNE~\cite{maaten2008visualizing} in order to provide an overview of their distribution and similarities (Fig.~\ref{fig:vis}, top part).  A pie chart glyph represents each of the topics. The colors in the glyph encode the word classes labeled by the experts which allows for more coarse coloring of the topics compared to the word level. By investigating and comparing the glyphs, experts can gain an interpretable information about the topics and assess their similarity. This part of the interface provides an overview that allow experts to investigate different granularities of the clustering, e.g., a small number of topics will lead to more general feature clusters, while a large number of topics helps to obtain a finer clustering. This interface also serves as an interactive panel in which experts can unite similar topics and by this compose informed clusters.

%
%The results of multiple LDA runs are collected into two matrices, the topic-word matrix and the sequence-topic matrix. %
The topic-word matrix is also displayed in the interface by means of a matrix visualization. The goal of the view is to enable an understanding of the topic features in terms of the word probabilities. %The x-axis shows the words, while the y-axis indicates the topics. Each cell represents the word probability in a specific topic, which is encoded by the opacity--the higher the opacity of a cell, the larger its probability in a specific topic. It is designed to support the identification of the keywords involved in the topics. The y-axis is sorted based on the similarity of the topics. Thus, the experts can easily capture the patterns of word distributions. 
%A Venn diagram visualizes the similarity between topics according to the sequence-topic matrix (Fig.~\ref{fig:vis} (c)). The circles represent topics, and their sizes indicate the number of associated sequences. The color encodes the word class that has the highest probability in the topic. Intersections between circles represent shared associated sequences. 
A chord diagram visualizes the similarity between topics according to the sequence-topic matrix (Fig.~\ref{fig:vis}, bottom part). The separate parts of the circle represent topics, and their sizes indicate the number of associated sequences. The color encodes the word class that has the highest probability in the topic. Interconnections between them represent shared associated sequences. 

Using the interactive interface human experts select a group of topics. The identified topic groups correspond to a partition of the data $X$ and are considered to be informed clusters. As a result of the visual analysis, we obtain $k\in\N$ clusters composed of topics $T$.
We propose the following method to perform inference of the cluster for a new sequence $s$: (i) produce probabilities for each of the topics in $T$ for $s$ (ii) identify the cluster of topics that has the largest average probability. Applying this scheme, we obtain a clustering $C$ that partitions the initial data into $\mathcal{G}_i \subseteq X$ and allows us to infer the cluster $C(s)$ of a novel sequence $s$. 
%, i.e., $\forall i,j\in [k]: \mathcal{G}_i\cap \mathcal{G}_j=\emptyset$ and $\bigcup_{i=1}^{k} \mathcal{G}_i = X$. Ideally, $k=m$ and the subsets each correspond to a different $\D_i$.
% When the domain experts are satisfied with their clustering, they can download the information describing the decomposition. These informed clusters of topics describe substructures in the existing normal data and are used to subdivide the input space. 
%
Due to the visual interface the clusters can be identified even by non-experts just visually and it does not require any effort, while still giving a benefit of easy sequences separation. We empirically verify the performance of informed clustering in the following section.
\section{Empirical Evaluation}
\label{sec:empirical_evaluation}
We evaluate the proposed novelty detection framework on three real-world datasets with diverse areas of application: cybersecurity, fake reviews, and server usage monitoring~\footnote{{The code is available at \href{https://github.com/kampmichael/noveltyDetectionSequentialData}{github}.}}. 
We show that per-cluster modeling is beneficial when compared to modelling an LSTM on all data (Global LSTM) and that informed clustering (IC-LSTMs) outperforms automatic clustering. Since the informed clustering is based on LDA, we use standard LDA as a baseline (LDA Cluster LSTMs), as well as standard k-means (k-means Cluster LSTMs). Note that for the cybersecurity dataset the informed clusters are produced by domain experts, while for the other two use cases clusters were produced by data scientists from our group. 
%This clustering is then fixed and used as decomposition for modeling with LSTMs.
% 
% We compare our results with informed clusters (IC-LSTMs) 
% %to the natural baseline of a single, global LSTM on all data. We also compare informed clustering 
% to automatic clustering techniques, i.e., k-means and LDA, in terms of the novelty detection performance. 
% For these automatic clustering methods it is necessary to set the number of clusters apriori. We optimize this parameter on the training set using silhouette analysis~\citep{rousseeuw1987silhouettes}. 
The LSTMs per cluster are trained with parameters obtained via a parameter evaluation on an independent subset of the data.
Furthermore, we compare per-cluster modelling to the natural baseline of using a clustering for outlier detection, in particualr the standard kNN with Minkowski distance and density-based clustering (HDBSCAN~\citep{mcinnes2017hdbscan}). We also compare to the standard outlier detection approaches one-class SVM (OC-SVM)~\citep{scholkopf2001estimating}, and isolation forests (IsoForests)~\citep{liu2008isolation}.
HDBSCAN, OC-SVM, and IsoForests are designed for tabular data and not sequences. Thus, we use the two common approaches for obtaining such features from sequences, the bag of words (BoW) approach, and using the top-800 sequence patters (SP) (obtained using the prefixspan algorithm~\citep{han2001prefixspan}), as well as the combination of both (BoW+SP). 
Hyper-parameter of HDBSCAN (min. cluster size, min. samples amount), OC-SVM ($\gamma$ and $\nu$ for RBF kernel), and isolation forests (max. number of features, number of estimators, and max. number of samples) are optimized on the test set (see published code for details), so that their results can be optimistic.
% Since the task of novelty detection on discrete sequences, where a whole sequence should be labeled as being anomaly, is rather specific, we select as baselines common outlier detection approaches, 
% %Furthermore, we compare the framework to general state-of-the-art approaches, 
% such as kNN, HDBSCAN~\citep{mcinnes2017hdbscan}, one-class SVM (OC-SVM)~\citep{scholkopf2001estimating}, and isolation forests (IsoForests)~\citep{liu2008isolation}. Each of these algorithms (except for kNN) requires a sequence being transformed into features in some way. 
% % %For kNN we use the original sequences without extracting any features. For the others we compare 
% We use bag of words features (BoW), the top-800 sequence patterns (SP) as features (obtained using the prefixspan algorithm~\citep{han2001prefixspan}), and the combination of both types of features. 
%
We report ROC AUC computed over all choices of the threshold $\theta$, and the sensitivity (Sens.) and specificity (Spec.) scores for the optimal value of $\theta$ in ROC space. We report mean of Sens. and Spec., i.e., a score that gives equal importance to both.
% We are interested in both sensitivity and specificity equally for outlier detection, since we do not want to miss outliers and we do not want to label normal instances as such. For this, we also report an average of these values, to understand which of the baselines performs best in both senses.
%

\subsection{Cybersecurity}
\begin{figure*}[ht]
\centering
\begin{minipage}[ht]{.4\textwidth}
  \centering
  \includegraphics[width=\linewidth]{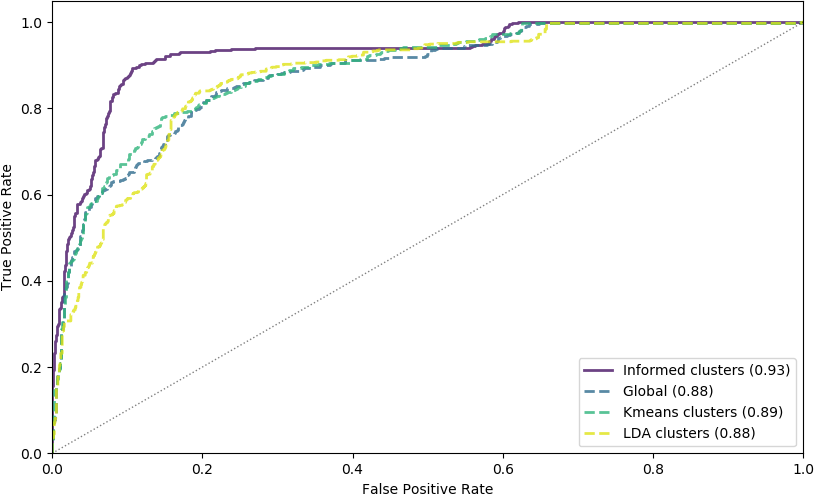}
  \caption{K-means, LDA and informed clustering (using LSTMs).}
  \label{fig:lstm_baselines}
\end{minipage}%
\qquad
\begin{minipage}[ht]{.4\textwidth}
  \centering
  \includegraphics[width=\linewidth]{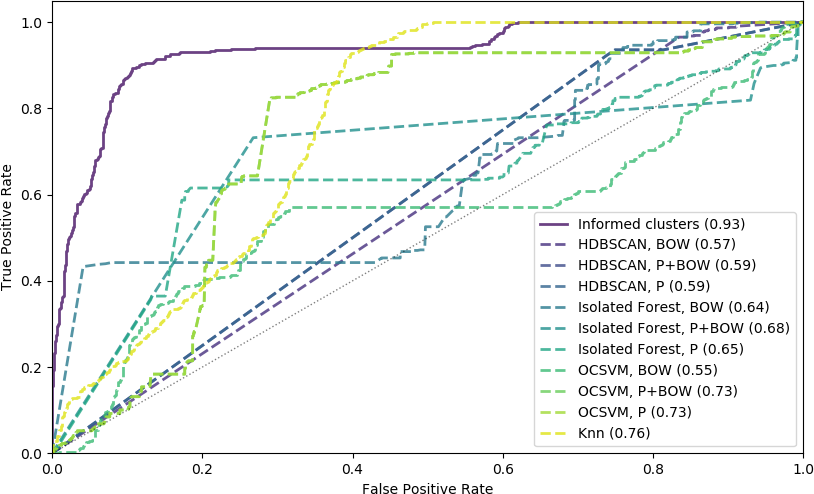}
  \caption{Anomaly detection baselines compared to the proposed approach.}
  \label{fig:baselines}
\end{minipage}
\end{figure*}
\begin{table}[ht]
    \setlength{\tabcolsep}{2pt}
    \centering
    \begin{tabular}{lrc|rr}
         \textbf{Method} & \textbf{AUC} & $\mathbf{\frac{Sens.+Spec.}{2}}$ & Sens. & Spec. \\
         \hline
         \textit{IC-LSTMs}                 & $\mathbf{0.93}$ & $\mathbf{0.89}$ & $0.89$ & $0.89$ \\
         Global LSTM                            & $0.88$ & $0.81$ & $0.80$ & $0.81$ \\
         k-means Cluster LSTMs                  & $0.89$ & $0.81$ & $0.83$ & $0.78$ \\
         LDA Cluster LSTMs                      & $0.88$ & $0.82$ & $0.81$ & $0.82$ \\\hdashline
         kNN with Mink. dist.            & $0.76$ & $0.61$ & $0.47$ & $0.74$ \\
         HDBSCAN on BoW                & $0.57$ & $0.77$ & $0.97$ & $0.57$ \\
         HDBSCAN on SP           & $0.59$ & $0.69$ & $0.70$ & $0.67$ \\
         HDBSCAN on BoW+SP            & $0.59$ & $0.69$ & $0.71$ & $0.67$ \\
         OC-SVM on BoW                 & $0.55$ & $0.5$ & $0.00$ & $0.99$ \\
         OC-SVM on SP            & $0.73$ & $0.5$ & $0.01$ & $0.99$ \\
         OC-SVM on BoW+SP             & $0.73$ & $0.5$ & $0.01$ & $0.99$ \\
         IsoForests on BoW       & $0.64$ & $0.49$ & $0.00$ & $0.97$ \\
         IsoForests on SP  & $0.65$ & $0.51$ & $0.05$ & $0.97$ \\
         IsoForests on BoW+SP   & $0.68$ & $0.54$ & $0.08$ & $0.99$ \\
    \end{tabular}
    \caption{Results for Cybersecurity Data}
    \label{tab:exp:resultsAUC_cybersecurity}
\end{table}

We first evaluate the proposed approach on a network intrusion detection task. We use the public ADFA-LD~\citep{creech2013generation} dataset. This dataset is considered to be the state-of-the-art benchmarking collection for evaluating intrusion detection approaches. The sequences of actions are logged from system calls in the Ubuntu Linux operating system.
%and it is specifically designed for anomaly-based systems. 
The normal behavior sequences are logged from usual activity, e.g., browsing through web pages or document editing. Attack sessions are generated according to known vulnerabilities of the system, e.g., brute force of user passwords. This application from cybersecurity is quite challenging, because attackers that want to infiltrate a network try to disguise their attacks as normal behavior~\citep{sommer2010outside}. Thus, accurately modeling normal behavior is crucial. 
Behavioral patterns were shown to provide important insights into the possibilities of attacks~\citep{ussath2017identifying, pannell2010user}. In this task, sequences are sessions of users accessing a system consisting of actions that they perform while being in this session. 

\begin{figure}[ht]
\centering
  \includegraphics[width=0.8\columnwidth]{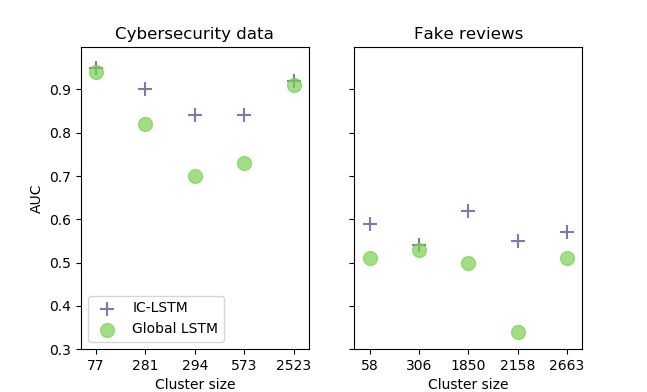}
  \caption{AUC per Cluster}
  \label{fig:exp:perClusterResults}
\end{figure}

%One or multiple consecutive unlikely actions constitute an anomaly and thus indicate malicious behavior. Since malicious activities has a tendency to disguise as normal ones, identifying them requires a precise model of normal behavior.
%However, learning global normal activity accurately without becoming too vague is often challenging due to the diversity of behaviors. A set of similar activities can be modeled more precisely~\citep{faraoun2006neural, chandrasekhar2014confederation}.
Using the visual analytics tool, we partition historical logs of activities into semantically sound behavior clusters. 
We obtained the clustering for our experiments in collaboration with security operators from an industrial partner.
%Since it is challenging for automatic clustering approaches to capture the semantics of a system~\citep{casas2012unsupervised}, our approach incorporates domain knowledge into the clustering process.

%\begin{table}[h]
%    \setlength{\tabcolsep}{6pt}
%    \centering
%    \begin{tabular}{p{0.1\textwidth}p{0.16\textwidth}p{0.13\textwidth}}
%         \textbf{Cluster Size} & \textbf{AUC IC-LSTM} & \textbf{AUC Global LSTM} \\
%         $2523$ & $\mathbf{0.92}$ & $0.91$ \\
%          $573$ & $\mathbf{0.84}$ & $0.73$ \\
%          $294$ & $\mathbf{0.84}$ & $0.70$ \\
%          $281$ & $\mathbf{0.90}$ & $0.82$ \\
%           $77$ & $\mathbf{0.95}$ & $0.94$ \\
%    \end{tabular}
%    \caption{Cybersecurity Data: Performance per Cluster}
%    \label{tab:exp:perClusterResults_cybersecurity}
%\end{table}
%

Table~\ref{tab:exp:resultsAUC_cybersecurity} shows the baseline results for the dataset. We report AUC for all the baselines together with mixture of sensitivity and specificity, as discussed in the introduction to the section. According to these two criteria our approach outperforms all the baselines. The ROC curve for all methods is shown in Figures~\ref{fig:lstm_baselines} and \ref{fig:baselines}. Our approach substantially outperforms baseline anomaly detection approaches, as well as a global model trained on all data. Moreover, the informed clustering performs better than automatic clustering using k-means or LDA. Thus, we conclude that LSTM-based methods are a suitable technique for this task and clustering before modeling is beneficial. Figure~\ref{fig:exp:perClusterResults} (left) shows the comparison of area under the ROC curve (AUC) per  cluster compared to the AUC of the global model on that cluster. Interestingly, the size of the cluster seems not to impact the outliers detection performance of the model. Rather, the performance of both local and global model are correlated over clusters. This suggests that the lower amount of training data has only little impact on the performance, whereas the quality of decomposition has a stronger effect.

\subsection{Fake Reviews}

\begin{table}[ht]
    \setlength{\tabcolsep}{2pt}
    \centering
    \begin{tabular}{lrc|rr}
         \textbf{Method} & \textbf{AUC} & $\mathbf{\frac{Sens.+Spec.}{2}}$ & Sens. & Spec. \\
         \hline
         \textit{IC-LSTMs}                 & $\mathbf{0.58}$ & $\mathbf{0.58}$ & $0.58$ & $0.57$ \\
         Global LSTM                            & $0.55$ & $0.54$ & $0.53$ & $0.55$ \\
         %k-means Cluster LSTMs                  & $0.58$ & $0.56$ & $0.57$ \\
         k-means Cluster LSTMs                  & $0.52$ & $0.51$ & $0.50$ & $0.52$ \\
         LDA Cluster LSTMs                      & $\mathbf{0.58}$ & $0.57$ & $0.57$ & $0.57$ \\\hdashline 
         kNN with Mink. dist.\footnote{Labels have been inverted.}            & $0.57$ & $0.56$ & $0.56$ & $0.55$ \\
         HDBSCAN on BoW                & $0.55$ & $0.5$ & $1.00$ & $0.00$ \\
         HDBSCAN on SP           & $0.56$ & $0.45$ & $0.45$ & $0.44$ \\
         HDBSCAN on BoW+SP            & $0.56$ & $0.45$ & $0.45$ & $0.44$  \\
         OC-SVM on BoW                 & $0.52$ & $0.5$ & $0.00$ & $0.99$ \\
         OC-SVM on SP            & $0.52$ & $0.5$ & $0.63$ & $0.36$ \\
         OC-SVM on BoW+SP             & $0.52$ & $0.5$ & $0.01$ & $0.99$ \\
         IsoForests on BoW       & $0.56$ & $0.5$ & $0.01$ & $0.99$ \\
         IsoForests on SP  & $0.54$ & $0.5$ & $0.01$ & $0.98$ \\
         IsoForests on BoW+SP   & $0.57$ & $0.49$ & $0.02$ & $0.96$ \\
    \end{tabular}
    \caption{Results for Fake Reviews Data}
    \label{tab:exp:resultsAUC_fakerevews}
\end{table}
The second task is identifying fake reviews. The dataset we use is collected from Yelp~\citep{mukherjee2013yelp}. The data includes reviews of hotels and restaurants in the Chicago area. Reviews include meta-information, as well as the text of the review itself, but for our experiments we use only the text. The labels are produced by a filtering algorithm that is used by Yelp. The labels are not perfect, but are proven to be sufficiently accurate~\citep{weise2011lie}. There are $13.23\%$ reviews evaluated as fake in this dataset. 
The texts were preprocessed using spacy (\url{https://spacy.io/}), i.e., lemmatized and having pronouns, punctuation and numbers replaced with uniform tokens. The overall vocabulary was cut to the frequency threshold $40$, meaning that all words that appear less than $40$ times are treated as unknown words. The amount of data was reduced for experiments in order to save preprocessing time, i.e., we use $10\%$ of the  $50000$ restaurant reviews.

%\begin{figure}[htb!]
%\centering
%  \includegraphics[width=1\columnwidth]{images/fake_reviews_per_cluster.png}
%  \caption{Fake Reviews Data: AUC per Cluster}
%  \label{fig:exp:perClusterResults_fakereviews}
%\end{figure}

The results in Table~\ref{tab:exp:resultsAUC_fakerevews} show an advantage of our approach, but not as pronounced as in the other tasks. It nevertheless outperforms all baselines, except for LDA Cluster LSTMs. Since the informed clustering uses LDA topics as well and LDA clustering is very successful in natural language processing, this is not surprising.
The k-means clustering LSTMs performs worse---even worse than the global LSTM---which is a further evidence that the clustering has to be chosen carefully.
Note that we have used the same one layered LSTM architecture with embeddings learned together with the task as in the other experiments. Thus, the results of the LSTM-based methods could be further improved by tuning the architecture.
Also note that the kNN applied on this data has a best AUC score lower than $0.5$, so the labels have been inverted (i.e., we treat the model as an oracle that is always mistaken).
The per cluster evaluation in Figure~\ref{fig:exp:perClusterResults} (right) shows that the proposed approach substantially outperforms the global LSTM on all clusters. This supports the idea that especially in vast and various data, a global model will become too general to notice the subtle deviations of novelties.

%\begin{table}[h]
%    \setlength{\tabcolsep}{6pt}
%    \centering
%    \begin{tabular}{p{0.1\textwidth}p{0.16\textwidth}p{0.13\textwidth}}
%         \textbf{Cluster Size} & \textbf{AUC IC-LSTM} & \textbf{AUC Global LSTM} \\
%            $2663$ & $\mathbf{0.57}$ & $0.51$ \\
%            $2158$ & $\mathbf{0.55}$ & $0.34$ \\
%            $1850$ & $\mathbf{0.62}$ & $0.50$ \\
%            $306$ & $\mathbf{0.54}$ & $0.53$ \\
%            $58$ & $\mathbf{0.59}$ & $0.51$ \\
%    \end{tabular}
%    \caption{Fake Reviews Data: Performance per Cluster}
%    \label{tab:exp:perClusterResults_fakereviews}
%\end{table}
%

\subsection{Time Series}

The third task is detecting novelties in real-valued time series. Since our approach is designed for discrete sequences, we need to discretize the time series by binning, and then cut them into non-overlapping windows. Each window that contains a time point marked as an anomaly is labeled as a novelty. All others are considered to be normal.
As an example of univariate real-valued time series we select data from the Numenta Anomaly Benchmark (NAB)~\citep{lavin2015evaluating}. NAB is a benchmark for evaluating algorithms for anomaly detection in streaming, real-time applications. It is composed of over $50$ labeled real-world and artificial time series data files plus a novel scoring mechanism designed for real-time applications. We considered the dataset of the AWS server metrics as collected by the AmazonCloudwatch service, in particular CPU Utilization numbers. Since the range of values in the sequences was from $0.062$ to $99.898$ we binned them uniformly into $1000$ bins from $0$ to $100$ with step $0.1$. The length of the window was chosen to be $40$ in order to have enough sequences in the training set.

%\begin{table}[h]
%    \setlength{\tabcolsep}{6pt}
%    \centering
%    \begin{tabular}{p{0.1\textwidth}p{0.16\textwidth}p{0.13\textwidth}}
%         \textbf{Cluster Size} & \textbf{AUC IC-LSTM} & \textbf{AUC Global LSTM} \\
%            $322$ & $0.97$ & $0.97$ \\
%            $137$ & $\mathbf{1.0}$ & $0.94$ \\
%            $108$ & $-$ & $-$ \\
%    \end{tabular}
%    \caption{Time series Data: Performance per Cluster}
%    \label{tab:exp:perClusterResults_timeseries}
%\end{table}
%
The results are displayed in Table~\ref{tab:exp:resultsAUC_timeseries}. Note that taking into account the simplicity of the task (because of the rather short sequences) all the LSTM based methods perform similarly. Also, due to having only $13$ anomalies to test on, the scores are very erratic---differing in $1$ example can change the AUC a lot. However, due to the small size, this dataset was the only one for which we can run kNN with the longest common sub-sequence (LCS) metric. The results obtained with it are $0.94$ AUC, sensitivity $0.85$ and specificity is $0.60$, but the hyper-parameter tuning alone took almost $120$ hours. As with all non-LSTM baselines, tuning was performed with test data, so it is optimistic. Again, the proposed IC-LSTM substantially outperforms all baselines; in general, LSTM-based approaches outperform the other baselines. Only kNN with Minkowski distance achieve competitive results. Note that the per-cluster comparison here is restricted to two clusters--the AUC of the last cluster could not be calculated, since it does not contain any novelties. These two clusters of size $322$ and $137$ have AUC $0.97$ and $1.0$ with our approach and $0.97$ and $0.94$ for the global LSTM model, indicating that the proposed approach is applicable to real-valued time series as well.

\begin{table}[htb]
    \setlength{\tabcolsep}{2pt}
    \centering
    \begin{tabular}{lrc|rr}
         \textbf{Method} & \textbf{AUC} & $\mathbf{\frac{Sens.+Spec.}{2}}$ & Sens. & Spec. \\
         \hline
         \textit{IC-LSTMs}                 & $\mathbf{0.99}$ & $\mathbf{0.87}$ & $0.77$ & $0.97$ \\
         Global LSTM                            & $0.96$ & $0.86$ & $0.77$ & $0.94$ \\
         k-means Cluster LSTMs                  & $0.97$ & $0.84$ & $0.85$ & $0.82$ \\
         LDA Cluster LSTMs                      & $0.98$ & $\mathbf{0.87}$ & $0.77$ & $0.97$ \\\hdashline
         kNN with Mink. dist.            & $0.97$ & $0.76$ & $1.00$ & $0.51$ \\
         HDBSCAN on BoW                & $0.78$ & $0.66$ & $0.31$ & $1.00$ \\
         HDBSCAN on SP           & $0.51$ & $0.31$ & $0.23$ & $0.39$ \\
         HDBSCAN on BoW+SP            & $0.52$ & $0.32$ & $0.23$ & $0.41$ \\
         OC-SVM on BoW                 & $0.85$ & $0.4$ & $0.08$ & $0.72$ \\
         OC-SVM on SP            & $0.79$ & $0.5$ & $1.00$ & $0.00$ \\
         OC-SVM on BoW+SP             & $0.79$ & $0.5$ & $1.00$ & $0.00$ \\
         IsoForests on BoW       & $0.55$ & $0.5$ & $0.00$ & $1.00$ \\
         IsoForests on SP  & $0.76$ & $0.25$ & $0.00$ & $0.49$ \\
         IsoForests on BoW+SP   & $0.76$ & $0.27$ & $0.08$ & $0.46$ \\
    \end{tabular}
    \caption{Results for Time series Data}
    \label{tab:exp:resultsAUC_timeseries}
\end{table}

\section{Conclusion}
\label{sec:conclusion}
The paper proposes a framework for novelty detection in discrete sequences combining decomposition and modeling. The empirical evaluation shows that decomposition improves the novelty detection accuracy substantially and that an informed clustering outperforms automatic ones on three different real-world datasets. 
%For discrete sequences, this approach outperforms state-of-the-art anomaly detection methods. This shows the importance and feasibility of decomposing complex tasks in unsupervised modeling. At the same time it shows that using knowledge from human experts is preferable to fully automatized methods.
Surprisingly, the models performed well even on smaller clusters where only little training data is available.
% In particular, we did not observe a significant drop in the performance of models on smaller clusters as it might be expected with decomposition approach and data-hungry models. 
A reason could be that slight overfitting due to a lack of training data is actually beneficial for novelty detection.
%Moreover, in the context of novelty detection, slight overfitting is less of a problem than in classification tasks.
%This could be due to smaller clusters representing simpler learning problems but the amount of data per cluster is too small to support this conclusion. 
Further studying the trade-off between decomposition and training set size is an interesting future task. 
In our experiments, domain experts were involved for the analysis of cybersecurity data, the reviews and server usage data was clustered by data scientists from our group. This is possible, since the visual interface is intuitive enough to perform meaningful clustering even without substantial domain knowledge. %While this slightly skews the results, we see this as an advantage of our methods. 
The strong performance of IC-LSTMs on the cybersecurity dataset indicates that using actual domain experts for clustering is still preferable.% further improve the other dataset results.
%But it also means, that even more accurate decomposition leads to even better results---cybersecurity dataset emphasized the performance of IC-LSTMs the most.

The proposed framework is designed for novelty detection in discrete sequences. 
%For future work, it is interesting to analyse the effect of decomposition on different data types, e.g., for classical tabular data, and whether informed clustering outperforms automatic clustering in those cases.
%
The flexible framework allows us to use a wide range of clustering and modelling techniques. In this paper, we restricted ourselves to LSTMs with the same architecture and parameters on all clusters, since they are well-suited for sequential data. An interesting future direction is to choose models for each cluster individually by tuning the models for an optimal bias-variance trade-off. 
% Of course, for neural networks the bias-variance trade-off is still poorly understood and even over-parametrized networks are capable of generalizing~\citep{zhang2016understanding}. Nonetheless, the framework allows for adapting the architecture of the neural networks for each cluster individually. Also, instead of using the same LSTM for each cluster, different neural networks (CNNs and RNNs) can be used as in~\cite{chawla2018host}. 

In this paper we assign a new sequence strictly to the closest cluster and use only the corresponding model for novelty detection. This approach could be improved by combining the predictions of multiple models. For most clustering techniques, including the informed clustering proposed in this paper, it is possible to infer a score for a new sequence that indicates the similarity to each cluster. This similarity score could be used to weigh the predictions of cluster models which might further improve the novelty detection accuracy. \cite{sharkey1999linear} shows that the combination of models trained on decomposed subtasks can be beneficial, but the way to combine predictions has to be selected carefully: opposed to ensemble techniques, the models might perform very poorly on other clusters and thus a simple averaging of predictions can be detrimental. Developing such combinations for our tasks is left for future work. 
Overall, we conjecture, that decomposition before modeling is beneficial for novelty detection in discrete sequences, on the condition that the clustering is of high quality and that often this can best be achieved with a knowledgeable human in the loop.

%Future work:
%- analysis of outliers proportions in the clusters when tested and how it affects results
%- finetuning instead of from scratch training?
%- other types of identifying suitable amount of clusters
%- effect of pretrained NLP embeddings

%\pagebreak
\bibliographystyle{icml2023}
%\scriptsize
\bibliography{bibliography}

\vfill
 	\pagebreak
	\appendix
    \onecolumn
	\section{Algorithm of the Framework}

\begin{algorithm}[H]
\SetAlgoLined
\KwResult{Set of informed clusters; LSTM for each of the clusters}
\caption{Training phase}
\begin{algorithmic}
 \STATE create input for the visual analytics tool
 \STATE present experts the interface for work
 \STATE download the set of informed clusters
 \STATE identify hyperparameters for LSTM on an independent subset
 \FOR{clusters in the set of informed clusters}
  \STATE train an LSTM
  \ENDFOR
\end{algorithmic}
\end{algorithm}

\begin{algorithm}[H]
\SetAlgoLined
\KwResult{Novelty score}
\caption{Inference phase}
\begin{algorithmic}
 \STATE get new sequence $s$
 \STATE infer informed cluster most probable for $s$
 \STATE calculate perplexity score from the model corresponding to it
\end{algorithmic}
\end{algorithm}

\end{document}